\documentclass[12pt]{iopart}
\usepackage[table]{xcolor}
\usepackage{graphicx}	
\usepackage[utf8]{inputenc} 
\usepackage[T1]{fontenc}    
\usepackage[colorlinks=true, 
            citecolor=blue,
            linkcolor=blue]{hyperref}       
\usepackage{url}            
\usepackage{booktabs}       
\usepackage{nicefrac}       
\usepackage{microtype}      

\usepackage{tcolorbox}
\usepackage{wrapfig}
\usepackage{orcidlink}
\renewcommand\appendixautorefname[1]{} 

\newcommand{\bx}{\mathbf{x}}
\newcommand{\bz}{\mathbf{z}}
\newcommand{\mathcolorbox}[2]{\colorbox{#1}{$\displaystyle #2$}}
\usepackage{iopams}  
\begin{document}

\title[Path Minimization for Latent ODEs]{Path-minimizing Latent ODEs for improved extrapolation and inference}

\author{%
  Matt L.~Sampson$^{\orcidlink{0000-0001-5748-5393}}$ \\ \address{Department of Astrophysical Sciences, Princeton University, Princeton, 08544, NJ, USA} \ead{matt.sampson@princeton.edu}
}

\author{%
  Peter~Melchior$^{\orcidlink{0000-0002-8873-5065}}$ \\ \address{Department of Astrophysical Sciences, Princeton University, Princeton, 08544, NJ, USA} \address{Center for Statistics and Machine Learning, Princeton University, Princeton, 08544, NJ, USA}\ead{peter.melchior@princeton.edu}
}


\begin{abstract}
Latent ODE models provide flexible descriptions of dynamic systems, but they can struggle with extrapolation and predicting complicated non-linear dynamics.
The latent ODE approach implicitly relies on encoders to identify unknown system parameters and initial conditions, whereas the evaluation times are known and directly provided to the ODE solver.
This dichotomy can be exploited by encouraging \emph{time-independent} latent representations. 
By replacing the common variational penalty in latent space with an $\ell_2$ penalty on the path length of each system, the models learn data representations that can easily be distinguished from those of systems with different configurations.
This results in faster training, smaller models, more accurate interpolation and long-time extrapolation compared to the baseline ODE models with GRU, RNN, and LSTM encoder/decoders on tests with damped harmonic oscillator, self-gravitating fluid, and predator-prey systems.
We also demonstrate superior results for simulation-based inference of the Lotka-Volterra parameters and initial conditions by using the latents as data summaries for a conditional normalizing flow.
Our change to the training loss is agnostic to the specific recognition network used by the decoder and can therefore easily be adopted by other latent ODE models. 
\end{abstract}

\begin{figure}
    \centering
    \includegraphics[width=0.99\textwidth]{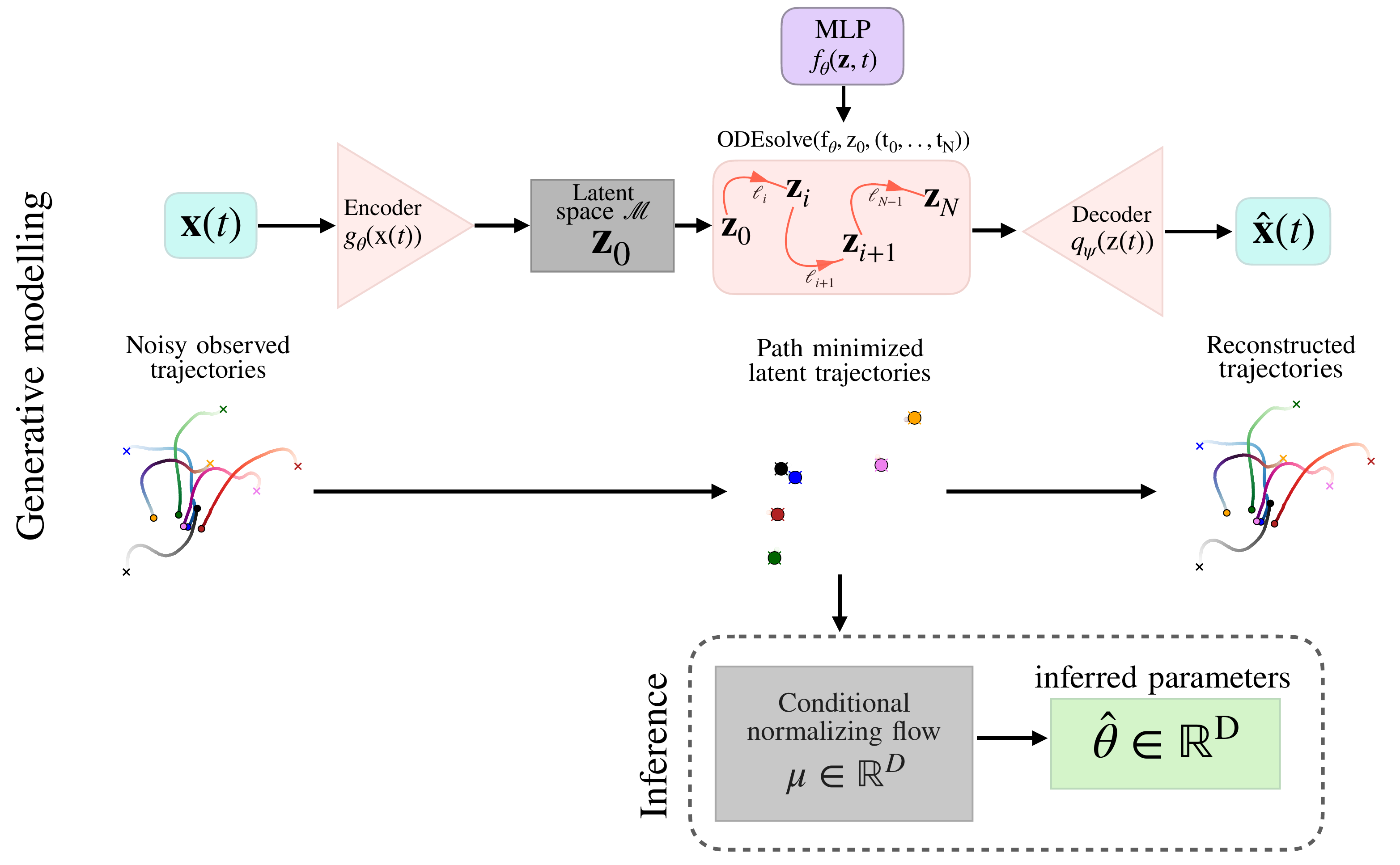}
    \caption{ {Schematic of our path-length minimizing latent ODE model.}}
    \label{fig:schematic}
\end{figure}

\section{Introduction}
Latent ODEs are a class of generative neural network models for sequential data, which have seen widespread adoption as flexible empirical descriptions of dynamic systems \cite{chen2018neural,rubanova2019latent}. A common architecture design uses a sequence model, such as a Recurrent Neural Network (RNN) \cite{Rumelhart1987} to initially encode an observed time series $\{\bx_1, \dots, \bx_n\}$. The latent representation $\bz\in\mathbb{R}^d$ is then evolved via numerical integration of the changes $\mathrm{d}\bz/\mathrm{d}t = f_\theta(\bz, t)$, represented by a neural network with parameters $\theta$, from the initial time $t_i$ to the endpoint $t_f$. The ODE is solved via a deterministic method, $\texttt{ODEsolve}(f_\theta, \bz_0, (t_0,\dots, t_N))$, allowing for estimates of $\bz$ at any arbitrary time-point $t\in [t_0, t_N]$, which can exceed the temporal range of the training data. Lastly, a decoder network predict the original signal $\bx(t)$.
It is customary to turn the neural ODE model into a Variational Autoencoder (VAE) \cite{VAE} to encourage a specific distribution in latent space, usually a $d$-dimensional Gaussian, but other distributions have also been shown to work well \cite{rubanova2019latent}.

While latent ODE models have been successful at reconstructing noisy and incomplete time series, they can struggle to learn long-term dependencies and thus to perform extrapolation \cite{lechner2020learning,xia2021heavy,shi2021segmenting, coelho2024enhancing, auzina2024modulated}. Part of this difficulty has been attributed to the vanishing/exploding gradient problem, leading \cite{lechner2020learning,coelho2024enhancing} to replace the RNN model with an LSTM model, but that was only moderately successful. \cite{auzina2024modulated} suggested that the issue lies in the mixing of static and dynamic state variables and proposed potential mechanisms for separating the two. We agree with this assessment and present a straightforward approach to establish this separation in arbitrary latent ODE architectures. 
Taking inspiration from differential geometry, where geodesics describe the shortest paths on a curved manifold, we posit that encouraging the autoencoder to form short paths in latent space effectively establishes such a manifold.
Doing so also drastically reduces the temporal variation of the latent representations.
We will show adding a path-length penalty to the loss improves simplicity and trainability of the models while producing more accurate predictions over a longer time frame.
We will also demonstrate that the latent representation allows for improved inference of system parameters and initial conditions by using them as data summaries in a simulation-based inference framework.

\subsection*{Key contributions}
\begin{itemize}
    \item We present a simple modification of the latent ODE loss function, replacing a variational KL penalty with an instance-based $\ell_2$ penalty in latent space, which can easily be introduced in current latent ODE models \cite{rubanova2019latent, shi2021segmenting, xia2021heavy, coelho2024enhancing, auzina2024modulated}
    \item We demonstrate that the latent space distribution becomes strongly shaped by parameters and initial conditions, which clarifies the interpretation of latent ODE encoders as inference models. This helps immediately whenever inference is the goal, and it establishes the design goal for suitable encoder architectures, e.g. to work on structured data sequences. 
 \end{itemize}

 The remainder of this paper is organized as follows: \autoref{sec:implementation} describes the implementation of the regularization method and relevant hyperparameters introduced. \autoref{sec:related} outlines related work in this area pointing out key differences between what has already been developed and what is presented here. \autoref{sec:experiments} presents the results from our  testing of this method against existing models and demonstrate the improved inference capabilities of this model. In \autoref{sec:discussion} we discuss limitations and future work in this area. We then summarise and conclude our findings in \autoref{sec:conclude}.

\section{Implementation}
\label{sec:implementation}
We adopt as baseline the architecture from \cite{rubanova2019latent}, which uses an ODE-RNN as the recognition network, trained as VAE. We also show trials with an ODE-GRU, and ODE-LSTM recognition network. We use a feed-forward neural network for the ODE function $f_{\theta}$ and the ODE solvers from the \texttt{diffrax} package \cite{kidger2021on} for numerical integration (full details in \autoref{sec:AppendixA}). However, as our primary interest lies in accurate forecasts and inference for given system configurations instead of fast sampling, we remove the variational aspect from the model, replacing the Gaussianity penalty with a path-length penalty:
\begin{equation}
\label{eqn:loss}
  \mathcal{L} =  \underbrace{ \sum_{i=1}^n (\bx_i - \tilde{\bx}_i)^2 }_{\mathrm{reconstruction \ loss}} \ + \underbrace{\mathcolorbox{blue!10!white}{ \lambda S }}_{\mathrm{min \ path \ loss}},
\end{equation}
where $\bx_i\in\mathbb{R}^f$ denotes an observation of the system at time $t_i$, $\tilde{\bx}_i$ its reconstruction by the latent ODE model, and $n$ the number of such observations.
The path-length penalty $S$ is given by the Mahalanobis distance between successive points $\bz_i$ and $\bz_{i+1}$ in latent space:
\begin{center}
\begin{tcolorbox}[width=13cm, colback=blue!8!white, colframe=white]
\begin{equation}
\label{eqn:path}
    S =  \sum_{i}^{m-1} \sqrt{(\bz_i - \bz_{i+1})^\top \mathbf{\Sigma}^{-1} (\bz_i - \bz_{i+1})}
\end{equation}
\end{tcolorbox}
\end{center}
The hyperparameter $m$ represents the number of interpolation points within the timespan of the observed trajectory (note that this can be much greater than $n$, the observed points),  and $\lambda$ controls the strength of the distance penalty. We search for best performance with a wide sweep, but we find good results for all tests with $\lambda \approx 1$ (see \autoref{sec:AppendixB} for more details). 

To prevent the latent distribution from collapsing to a point, which trivially satisfies the path-length penalty, we scale the penalty with the empirical standard deviation in latent space $\mathbf{\Sigma}$ of batches of trajectories from different system configurations. This diagonal matrix gets updated after every training step. 

We point out here that while the observed data points $\{\bx_1,\dots,\bx_n\}$ may be sparse, we can sample latent points for any arbitrary time $t$. The hyperparameter $m$ effectively controls the accuracy of the discrete interpolation of the trajectory in the latent space, making this method suitable even in the presence of large temporal gaps in the data. But one must ensure that the temporal differences between successive $z_i$ values is at least as large as the integration time step $dt$ due to numerical constraints of the solver.

\section{Related work}
\label{sec:related}
\subsection{Regularized neural ODEs}
The idea of including a regularizer for neural ODEs has been explored before \cite{finlay2020train,ghosh2020steer, kelly2020learning, lee2023minimizing}. These methods primarily invoke concepts from optimal transport \cite{behrmann2019invertible}, with regularization terms resulting in straight-line trajectories in latent representation spaces. The purpose is to reduce integration times for the numerical solver due to larger possible time steps. Doing so has been shown to improve training times but not to significantly affect accuracy at either long or short time reconstructions of the trajectories. These methods may even require the computation of higher-order derivatives, negating some of the aforementioned speed advantages \cite{kelly2020learning}.

Our approach, while still a form of regularization, differs in that we place no constraints on the shape or curvature of the path in latent space. 
We simply require short paths.
This is more akin to searching for a geodesic (often highly curved) solution of a dynamic system as opposed to enforcing straightness, which does not appear necessary or beneficial for the integrator.

\subsection{Long-term dependencies} 
Learning the long-term dependencies of dynamical systems is an important goal when studying sequential data. Part of the problem with neural ODE models is the vanishing gradient problem arising from the use of RNNs in the ODE encoder. \cite{lechner2020learning} and \cite{coelho2024enhancing} present neural ODE and latent ODE models, respectively, that use a long-short-term-memory (LSTM) model \cite{schmidhuber1997long} as the encoder to help mitigate this problem. An alternate approach, HBNODE, is taken by \cite{xia2021heavy}, who use the \textit{heavy ball} method \cite{polyak1964some} as a way of stabilizing the gradients from very long sequences, which results in better extrapolation. Our approach is similar in that it changes the training, albeit with conventional gradient descent, so it can be adopted with any encoder/decoder architecture.

\subsection{Modulated neural ODEs}
\cite{auzina2024modulated} introduce a novel modulation scheme for neural ODE (MONODE), which separates the static, dynamic, and latent state variables. This is done through an explicit forcing of time invariance on the static and dynamic \textit{modulator} variables, which are estimated through an separate encoder networks trained simultaneously with the rest of the model. This results in improved extrapolation accuracy compared to more traditional neural (latent) ODE methods. However, it requires an a priori splitting of the latent space and still requires a variational penalty on the latent space, whereas our modification aims to achieve a similar goal without changes to the latent ODE architecture. The authors also note poor out-of-domain performance of their models. 

\section{Experiments}
\label{sec:experiments}
We carry out three test cases of increasing complexity and degrees of freedom, with periodic and non-periodic solutions. Specifically, we test both models on a damped harmonic oscillator (periodic), a self-gravitating fluid (non-periodic), and a predator prey system (periodic). As our main \emph{baseline} we adopt the latent ODE-RNN architecture described in \cite{rubanova2019latent} as well as ODE-GRU for all tests and ODE-LSTM for the harmonic oscillator test. We omit showing results from the ODE-LSTM recognition network for the latter two test cases due to training difficulties of the model. For our most complicated task (predator--prey) we compare our model against the alternative training method HBNODE. We record the training iterations, interpolation accuracy, and the extrapolation accuracy of our model in all cases. We use 80$\%$ of each dataset for training, 10$\%$ for testing, and 10$\%$ for validation. Full details about data generation are in \autoref{sec:AppendixA}, while training, hyper-parameters, and model parameters for all test cases are detailed in \autoref{sec:AppendixB}. 

Importantly, we limit the size of recognition networks in our models, with the size of the recognition network at least two or three times smaller than that used for similar tests in similar studies \cite{chen2018neural,rubanova2019latent,shi2021segmenting,coelho2024enhancing, auzina2024modulated}. We found increasing the recognition model size can lead to overfitting the interpolated region at the cost of the extrapolated regions, particularly in the baseline model. Using large model sizes for the baseline would significantly increase training time for no guarantee of better extrapolation performance\footnote{See \autoref{sec:AppendixB} for details on this}. We compare our results on recognition networks of the same size, with the exception of the HBNODE trials for the predator--prey system. For all trials, we allow the baseline model to train for up to 10 times as long as the best performing path-minimized model and report results for the best trained model.

\subsection{Damped harmonic oscillator}
\label{sec:dho}

\begin{figure}[t]
    \centering
    \includegraphics[width=0.99\textwidth]{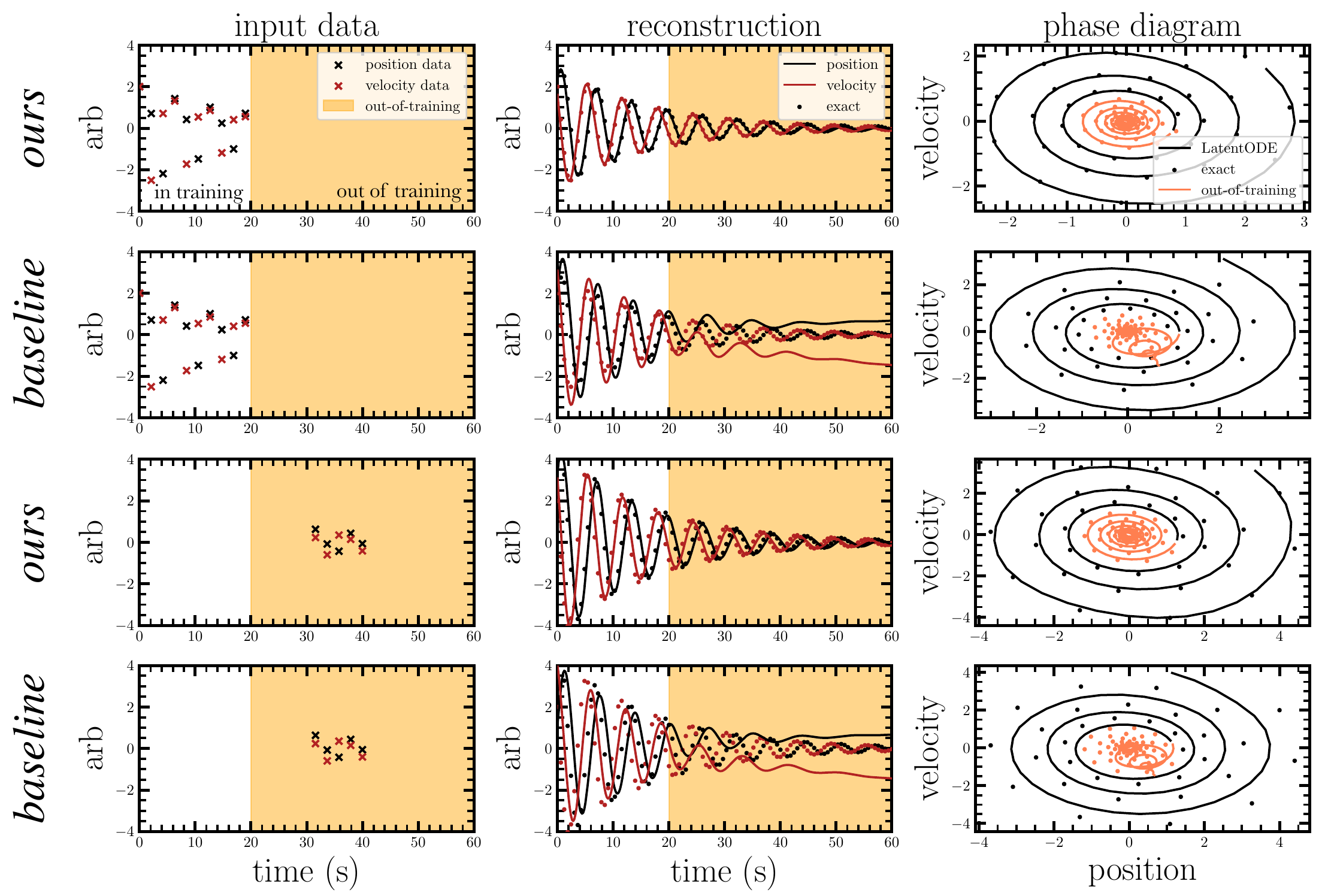}
    \caption{ {\emph{Column one}: Test data (black for position, red for velocity) for the damped harmonic oscillator model. Times beyond those encountered during training are shaded in orange. \emph{Column two}: The results of the model reconstruction. The line plots indicate the predictions from the latent ODE-RNN models and the dots the exact solutions. \emph{Column three}: A phase space plot with lines from the reconstruction and dots from the exact solutions. Orange color again indicates the times where the model has to extrapolate post the training regime. We note that our model was trained for 5,000 training steps, while the baseline was trained for 10,000 steps.}} 
    \label{fig:sho_plot}
\end{figure}

Our first test case is run on a damped harmonic oscillator (DHO, details of data generation are in \autoref{sec:AppendixA}). In the first column of \autoref{fig:sho_plot} we show two examples from the test set, for which we randomly choose initial conditions, randomly choose observation times from a uniform distribution between $t_0$ and $t_f$ to yield $n=20$ and $n=5$, respectively, and add Gaussian jitter to all the points.

We show predictions for both the velocity and the position of the trained models in the second column of \autoref{fig:sho_plot}. The third column shows the same reconstructions in a phase space plot with the orange colored parts of the lines indicating extrapolated regions.
Our path-minimizing loss function results in more accurate reconstructions in both the interpolation and extrapolation regime (the latter refers to times not encountered during training, shaded in orange). The baseline model shows good, although not excellent, performance for interpolation, but fails to capture long-time extrapolation features.

\begin{figure}
    \centering
    \includegraphics[width=0.92\textwidth]{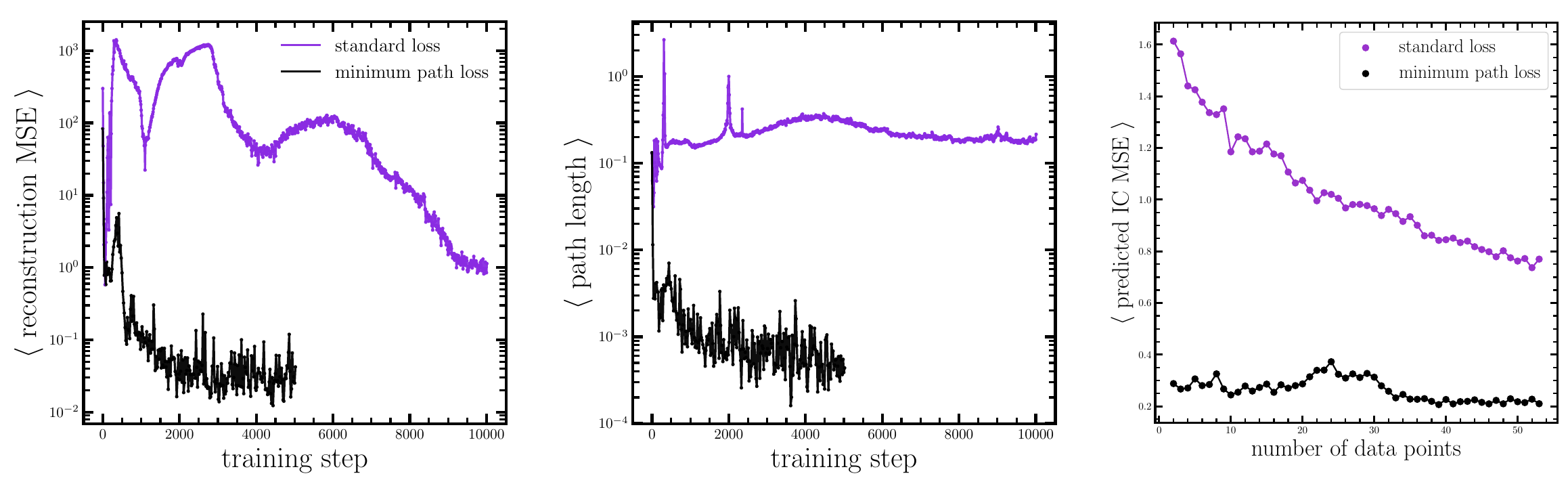}
    \caption{ {The reconstruction error for extrapolated trajectories (\emph{left}) and the path length calculated from \autoref{eqn:path} (\emph{center}) as function of training iteration for the baseline (purple) and our path-minimizing model (black). \emph{Right:} The error in predicting the initial conditions of the DHO as a function of data points given for modelling. We uniformly sample $n$ data-points between $t=5$ and $t=60$ adding Gaussian random noise to each point. All results are averaged over 256 trials.}} 
    \label{fig:dho_metrics}
\end{figure}

\autoref{fig:dho_metrics} shows the MSE of the reconstructed dynamics for both extrapolation and interpolation (left panel),the average path length as calculated via \autoref{eqn:path} taken by the integrator in latent space (center panel), and the MSE of inferring the initial conditions (ICs) by running the integrator to $t=0$ from inputs with a range of sizes (right panel). All quantities are averaged over the batch size of 256 of each training step. 
Our model trains much more effectively, can easily improve  reconstruction loss and path-length loss at the same time, and yields more accurate results at any time, including $t=0$. A more detailed summary of the results is given in \autoref{tab:results} with additional results shown in \autoref{sec:AppendixB}.

\subsection{Lane-Emden equation}
\label{sec:lane}

Our second test is on the Lane-Emden equation, which describes a symmetric self-gravitating fluid:
\begin{equation}
    \label{eqn:lane}
    \frac{1}{\xi^2} \frac{d}{d\xi} \left(\xi^2 \frac{d\theta}{d\xi} \right) + \theta^n = 0,
\end{equation}
where $\xi$ is a dimensionless length radius, $\theta$ describes the fluid density and pressure, and $n$ is the polytropic index of the fluid. We note that while $n$ can theoretically be any real numbered value between $0 \to \infty$, we limit our sampling to $n \in \{1, 2, 3, 4, 5\}$ as outside of these values strong constraints need to be placed on the radius $\xi$ for numerical stability \cite{vanani2010numerical,baty2023modelling}. There are fixed initial conditions for this problem with $\theta_0 = 1$, $\frac{d}{d\theta_0} = 0$, hence there is 1 degree of freedom in the equation. 

\begin{wrapfigure}[26]{r}{0.43\linewidth}
    \includegraphics[width=\linewidth]{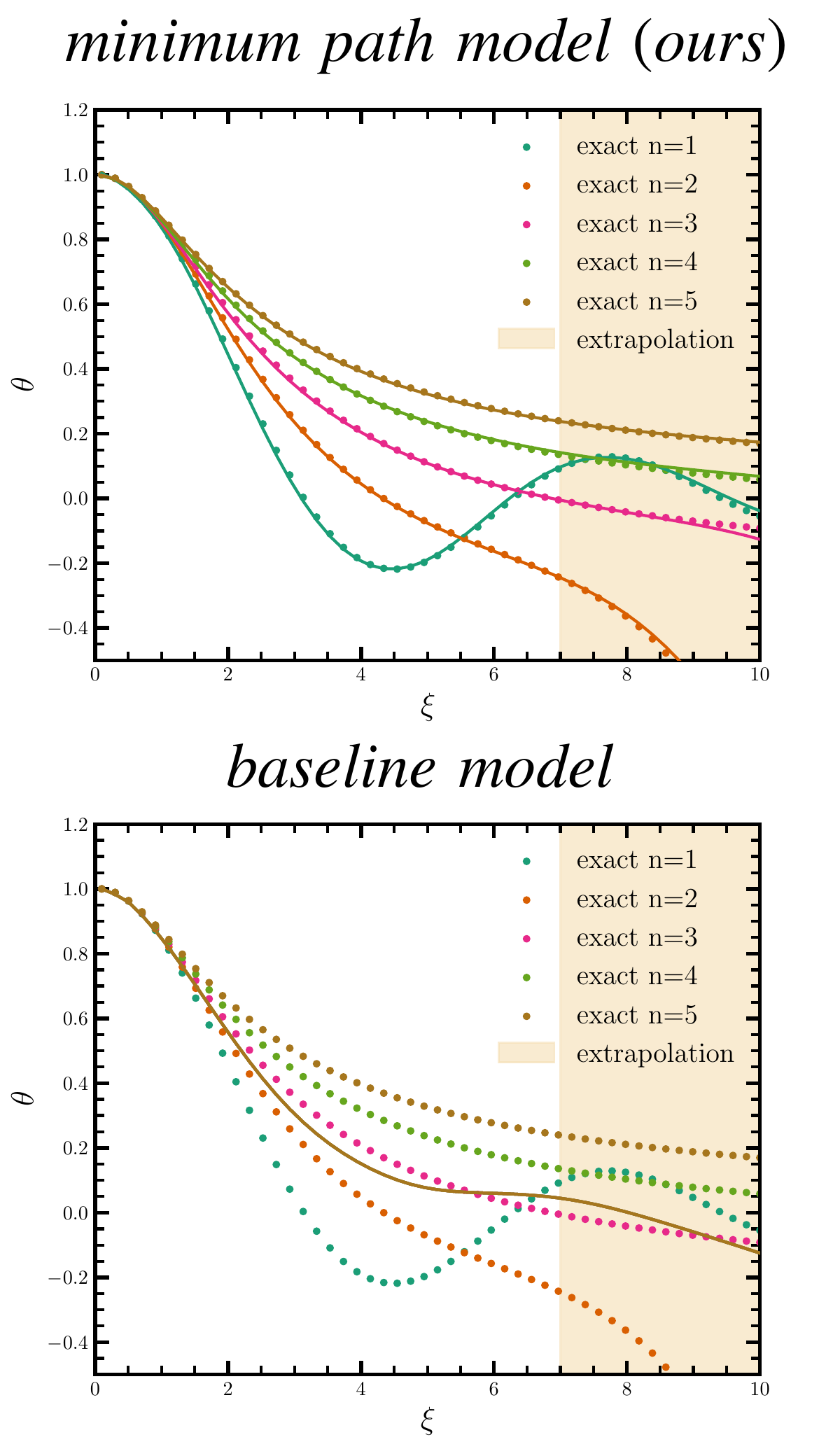}
    \caption{{Latent ODE reconstructions (lines) vs numerically exact solutions (dots) to \autoref{eqn:lane} for integer polytropic indices $n$.}}
    \label{fig:lane}
\end{wrapfigure}

\autoref{fig:lane} shows the latent ODE reconstruction (lines) vs the numerically exact solution (dots) for 5 different polytropic indices. We exclude the first $n=0$ mode as this rapidly diverges to 0, making long-term extrapolation infeasible.

We perform a large variety of trials with different model sizes, encoder dimensions, and training routines for the baseline model, which we detail in \autoref{sec:AppendixB}, but we were unable to get any reasonable solutions for this test.
This could be due to the discrete nature of our solution space, which increases the difficulty of searching for a single valid solution to all 5 values of $n$. The results demonstrate the significantly improved performance of our model on a discrete\footnote{Effectively discrete because we follow the standard practice of restricting $n \in \mathbb{N}$ for this problem.}, non-periodic ODE. However, our path minimizing  ODE-RNN model is capable of generating accurate solutions for non-integer values, which it has not seen during training (see details in \autoref{sec:AppendixC}).

\subsection{Lotka-Volterra equations}
\label{sec:lve}
We also test on the Lotka-Volterra equations (LVE), a nonlinear set of coupled first-order ODEs 
\begin{equation}
    \label{eqn:lve}
    \frac{dx}{dt} = \alpha x - \beta x y, \ \frac{dy}{dt} = \delta x y - \gamma y,
\end{equation}
that is often used to describe the interdependence of the numbers of predators $x$ and their prey $y$.
The LVEs have been found to present difficulties for latent ODE type models \cite{shi2021segmenting,auzina2024modulated}, especially their long-term behavior, and therefore present a good test case for our purposes. 

\begin{figure}
    \centering
    \includegraphics[width=0.98\textwidth]{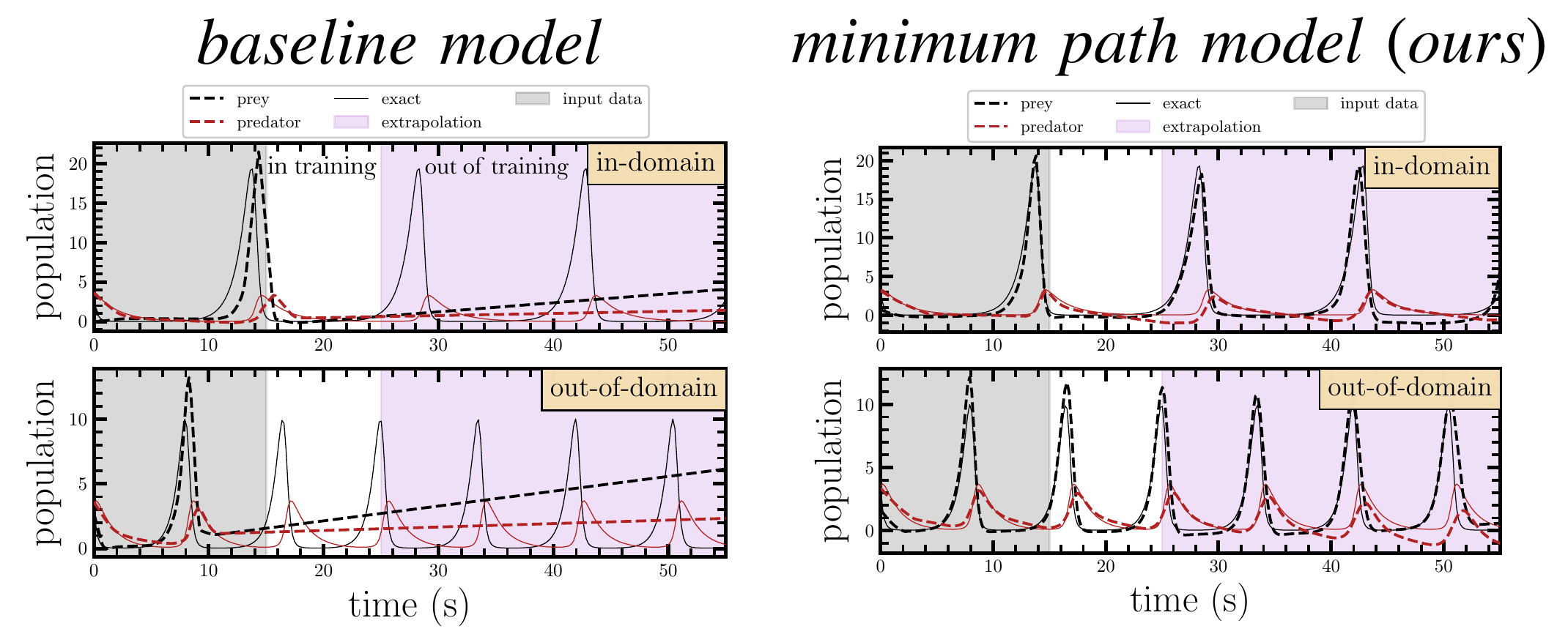}
    \caption{ {Reconstructions of solutions for the Lotka-Volterra equations with randomly sampled initial conditions and model parameters from within the training ranges (\emph{top}) and from up to 25\% beyond the training ranges (\emph{bottom}). The grey shading indicates the region where data is supplied, the purple shading indicates extrapolation regions, i.e no model has seen training data past this point.}} 
    \label{fig:lve_long}
    \vspace{-1em}
\end{figure}

We use sample uniformly parameters $\alpha \in [1.0 , 3.5], \; \beta \in [1.0, 3.5], \; \delta \in [0.5, 0.6], \; \gamma \in [0.5, 0.6 ]$ and initial conditions $y_0 \in [1, 6]$ as in \cite{auzina2024modulated}. We plot two example trajectories, one \textit{in-domain} and one \textit{out-of-domain} with their analytic solution in \autoref{fig:lve_long}. For the out-of-domain data, we move each of the 4 parameters $25\%$ outside of their respective training ranges. The baseline model fares well in reconstructing in the interpolation regions ($t=0 \to 25$), with increasing errors at longer times, for the in-domain test. The performance drops drastically for the out-of-domain test, especially past the point where we supply data (grey shading). The extrapolated dynamics, indicated by the pink shaded regions of \autoref{fig:lve_long}, are almost straight extensions, which is implausible for predator-pray systems. These results are in agreement with \cite{shi2021segmenting}, who also find poor extrapolation performance of baseline model on the LVE.

In contrast, our model on the right shows high accuracy even up to large times, showing the expected cyclic behavior of the LVE. It maintains high levels of accuracy even for out-of-domain testing. We compare our results for this test against HBNODE, which performs better during interpolation but substantially worse in extrapolation. Summary statistics are listed in \autoref{tab:results}.

\begin{table}
    \centering
    \caption{Test data MSE and its standard deviation (averaged over 256 trials) for damped harmonic oscillator and predator-prey system of our model and the baseline encoder models. We show results for interpolation ($t=20, t=25$) and extrapolation ($t=90, t=50$). We note the HBNODE trials were run with altered sequential training schemes.}
    \begin{tabular}{cccccc}
    \toprule
       Test case  & encoder & $t_{final}$ &  Baseline model & Minimum path (ours) \\ \midrule
          &  &   & $\langle$ MSE $\rangle$ (std) & $\langle$ MSE $\rangle$ (std) \\ \midrule
        Damped oscillator & ODE-RNN & 20 &  24.59 (0.03) & \textbf{0.328 (0.006)} \\ 
         & ODE-LSTM & 20 &  48.65 (0.371) & 0.532 (0.153) \\
         & ODE-GRU & 20 &  5.48 (0.283) & 0.827 (0.121) \\   \arrayrulecolor{black!30}\midrule
         & ODE-RNN & 90 &  730.8 (2.79) & \textbf{0.767 (0.01)}\\
         & ODE-LSTM & 90 & 185.2 (0.46) & 68.86 (0.212) \\
         & ODE-GRU & 90 &  6632 (12.37) & 1.599 (0.231) \\   \arrayrulecolor{black!99}\midrule
         Lane Emden & ODE-RNN & 7 &  0.064 (0.06) & \textbf{0.0011 (0.005)} \\ 
         & ODE-GRU & 7 &  0.066 (0.07) & 0.0022 (0.05) \\   \arrayrulecolor{black!30}\midrule
         & ODE-RNN & 10 &  0.118 (0.17) & \textbf{0.004 (0.012)} \\
         & ODE-GRU & 10 &  0.125 (0.19) & 0.018 (0.10) \\   \arrayrulecolor{black!99}\midrule
        Predator-prey  & ODE-RNN
        & 25 &  13.96 (0.57) & 5.19 (0.32) \\ 
           & ODE-GRU & 25 &  16.16 (4.12) & 12.18 (2.92) \\   & HBNODE & 25 &  \textbf{2.851 (0.49)} & N/A \\
          \arrayrulecolor{black!30}\midrule
           & ODE-RNN & 50 &  27.87 (0.86) & \textbf{12.507 (0.54)} \\
         & ODE-GRU& 50 &  17.49 (3.77) & 13.06 (2.71) \\   
         &  HBNODE &  50 & 440.6 (24.6) & N/A \\ \arrayrulecolor{black!99}\bottomrule
    \end{tabular}
    \label{tab:results}
    
\end{table}



\subsection{Parameter inference}
\label{sec:inference}

\begin{wrapfigure}[20]{r}{0.45\linewidth}
\vspace{-4em}
    \includegraphics[width=0.98\linewidth]{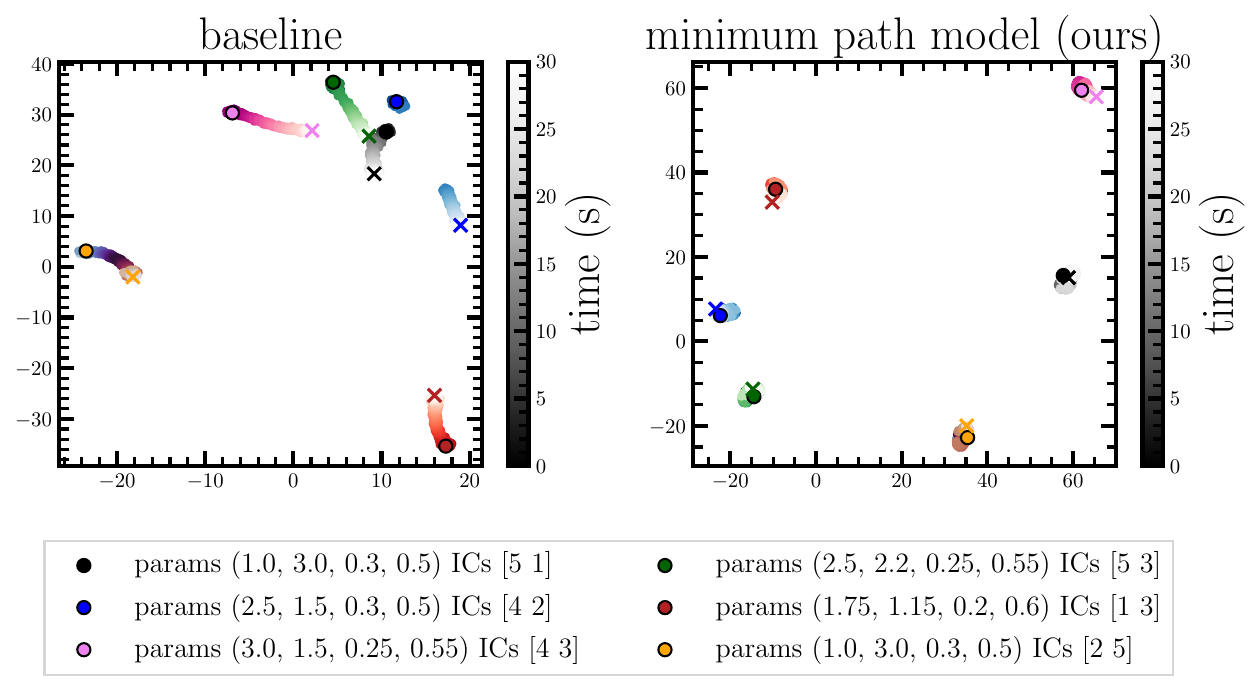}\\    
    \includegraphics[width=0.98\linewidth]{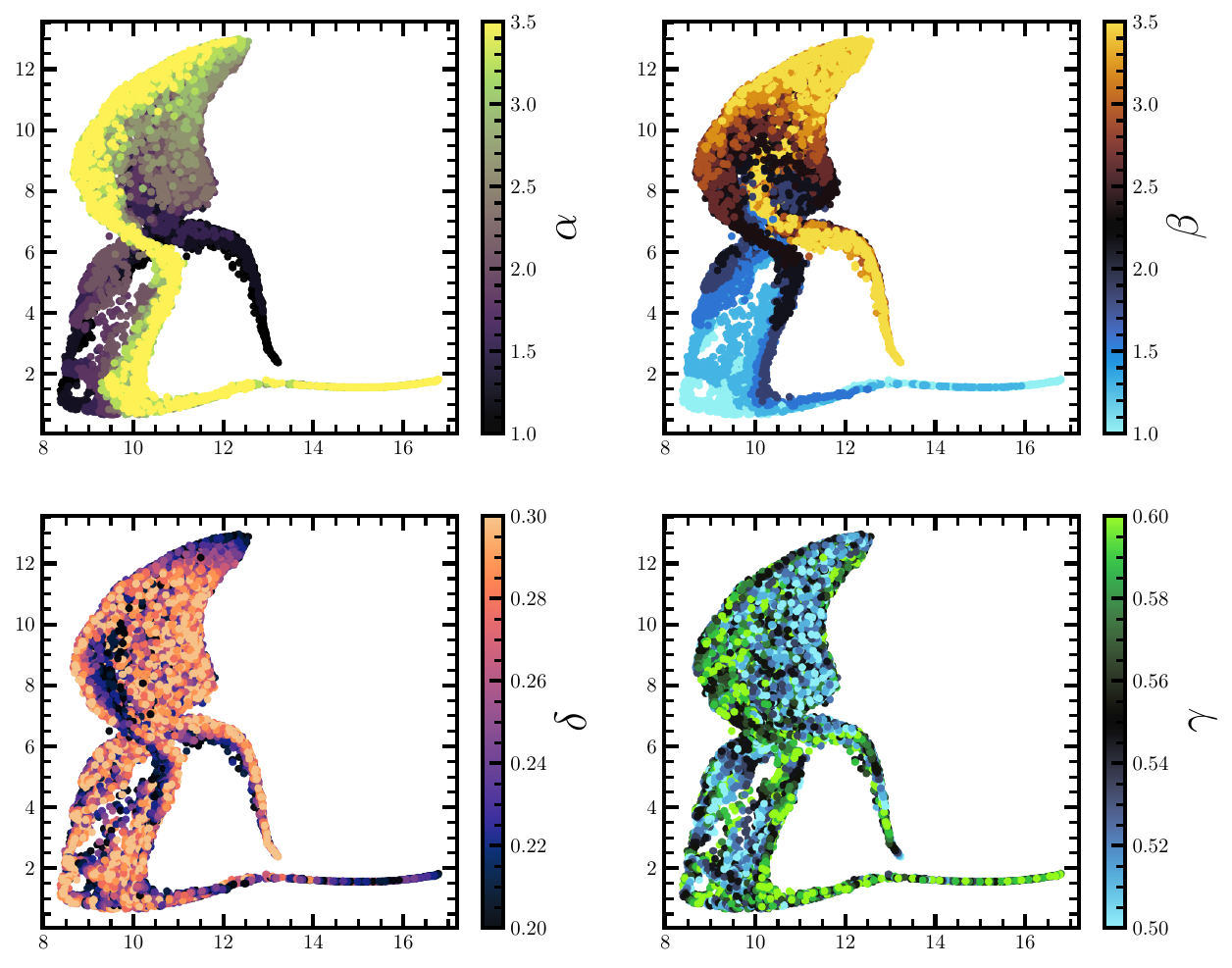}
    \caption{UMAP projection of six different latent space trajectories for the LVE (\emph{top}) and of 10,000 different LVE parameters for the path-minimizing model (\emph{bottom}). 
    }
    \label{fig:lve_umap}
\end{wrapfigure}

A relatively unexplored avenue is the use of latent ODEs as inference models. \cite{xia2021heavy,auzina2024modulated} discuss it in relation to long range-extrapolation and static/dynamic variable splitting, respectively. However, limited results or progress has been shown at the time of writing. 
Our modified loss does not require Gaussianity in latent space of the baseline with its KL penalty, instead, for every set of observations of a specific dynamic system, it penalizes distances in latent space as a function of time. As the center panel of \autoref{fig:dho_metrics} shows, the loss effectively shrinks average path length for DHO systems.

The top panel of \autoref{fig:lve_umap} make this point more clearly. 
We show a UMAP \cite{mcinnes2018umap} projection of the LVE trajectories in latent space for a selection of trials (shown in different colours). As expected, the addition of a path-length penalty shortens the trajectories in our model compared to the baseline by an average factor of $\approx30$.
As a result, despite the time-dependent nature of dynamical systems, the ODE integrator only fills small regions of latent space, leaving most of its volume free to capture the time-invariant descriptors of the system: parameters and initial conditions. 

This behavior becomes clear when we fill the UMAP with latents from our model for 10,000 randomly selected LVE system parameters and color-code by the parameter values. The bottom panels of \autoref{fig:lve_umap} show evident structure, which means that specific parameters are located in specific locations in latent space, and that variations of parameters leads to smooth changes of latent space positions. This last feature may provide some robustness when parameters are sampled from out-of-distribution of the training data.

The evident structuring of the latent space to follow primarily parameter and IC values should allow for accurate inference from noisy and irregularly sampled observations even without the explicit splitting of static and dynamic variables proposed by \cite{auzina2024modulated}. 
The right panel of \autoref{fig:dho_metrics} already presented the error in reconstructing the initial conditions for the DHO. It shows that our model recovers the ICs better even with as low as  $n=3$ observed samples than the baseline model with $n=50$. But this finding depends on the information content of the latents and the accuracy of the ODE integrators.

To more concretely test the inference capabilities of our model, we train a normalizing flow \cite{rezende2015variational} to predict six effective parameters\footnote{There are four parameters in \autoref{eqn:lve} plus the initial value of predator and prey numbers.} of \autoref{eqn:lve} given the latent context vector from the autoencoder as an input \cite{Winkler2019}. In the language of simulation-based inference, we use the latent ODE encoder to produce summaries for potentially irregularly sampled sequence data.
\begin{figure*}
    \includegraphics[width=0.98\linewidth]{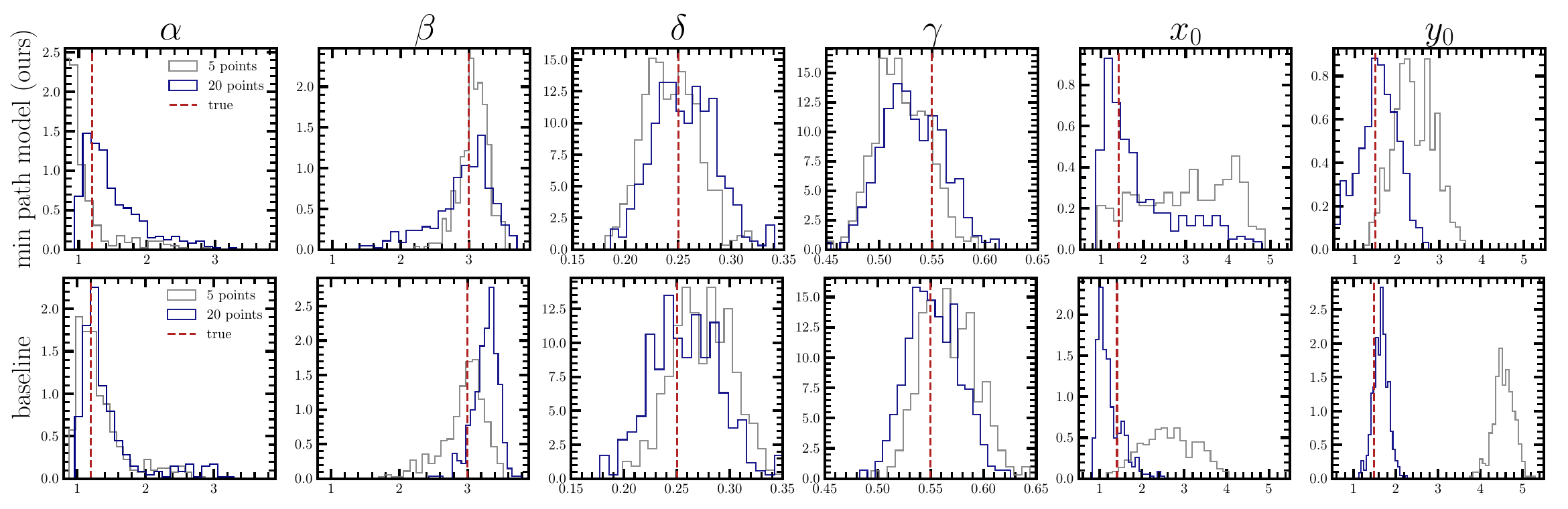}
    \caption{ {Posterior plots from an LVE test case using a normalizing flow to infer the value of the parmeters and initial conditions, for our model (\emph{top}) and the baseline model (\emph{bottom}). The gray histogram shows the posteriors for $n=5$, the blue for $n=20$ observations. The red dashed line indicates the true parameter value.}} 
    \label{fig:corners}
\end{figure*}
We show a sample posterior plot in \autoref{fig:corners} for the inference of a fixed set of parameters and ICs for a visual comparison of the model performance. The top panels show the results for our model, the bottom panels for the baseline. We plot results for inference made with 5 and 20 randomly sampled observations from the interpolation regime in gray and blue, respectively, with the true values indicated by vertical dashed lines.
We can see that baseline and path-minimizing model produce latents that are inherently useful for inference. This can partially be attributed to the supervised training of the normalizing flow, which can tolerate substantial confusion of the inputs (such as the crossing of latent trajectories seen in the top panel of \autoref{fig:lve_umap}) and still produce reasonable outputs.
On closer inspection, we see that providing more data leads to narrower posteriors, as expected from an inference model.
However, in the $n=5$ case the baseline model is heavily biased in the inference of the initial conditions.

We repeated these experiments by uniformly sampling both parameters and initial conditions then taking 5, 10, and 20 data points samples between $t=0$ and $t=15$ and reporting the relative MSE of the parameter and IC posterior means averaged over 256 trials. The results are summarized in \autoref{tab:inference}. We test the inference capabilities for in-domain and out-of-domain parameter sets, where for the out-of-domain set we extend the parameter bounds to exceed the training range of the latent ODE model by $50 \%$ on both the high and low ends.

\begin{table}[t!]
    \centering
    \caption{ Inference accuracy (relative MSE of the posterior mean and its standard deviation computed over 256 trials) for the LVE parameters.}
    \begin{tabular}{cccc}
    \toprule
       num points &  out-of-distribution & Baseline model & Minimum path (ours) \\ \midrule
          &   & $\langle$ rel. MSE $\rangle$ (std) & $\langle$ rel. MSE $\rangle$ (std) \\ \midrule
        5 &  No & 1.879 (1.88) & \textbf{0.320 (0.22)} \\ 
        10 &  No & 0.708 (0.72) & \textbf{0.261 (0.24)} \\
        20  &  No & 0.301 (0.30) & \textbf{0.197 (0.20)} \\ \midrule
        5 &  Yes & 2.045 (1.97) & \textbf{0.482 (0.31)} \\ 
        10 &  Yes & 0.930 (1.2) & \textbf{0.406 (0.34)} \\
        20  &  Yes & 0.464 (0.39) & \textbf{0.347 (0.25)} \\ \bottomrule
    \end{tabular}
    \label{tab:inference}
    
\end{table}

Our model consistently outperforms the baseline in all trials, with notably stronger performance for sparsely observed data. In fact, the posterior MSE of our model with $n=5$ is comparable to the baseline model with $n=20$.
This behavior is retained for out-of-distribution cases.

\section{Discussion and future work}
\label{sec:discussion}
\subsection{Complexity of the ODE} In our experiments we find a strong relation between the complexity of the underlying ODE and the benefits from our path-minimizing loss. The simplest case of the damped harmonic oscillator is where the baseline model performed the strongest. We see significant shortcomings for either more complex systems and/or ones with more degrees of freedom as found by the self-gravitating fluid and the predator-prey systems. While it is feasible that significantly increasing the size of the ODE network itself may help in these cases, this raises concerns about overfitting to interpolated regions, which is likely worsening the performance in extrapolation and reducing the overall robustness of the latent ODE model.

\subsection{Data sparsity}
We find significant differences between the performance of the baseline and our model in situations of sparsely observed sequences. The path-minimizing model is robust in interpolation, extrapolation, and parameter inference, even when the number of supplied data points is significantly smaller than that used during training. We attribute this robustness to a more effective training of the encoder if the ODE solver produces short trajectories on low-dimensional submanifolds, namely those with information about system configuration. The encoder thus benefits from better-defined targets, which renders it feasible to predict meaningful latents from sparse inputs.

\subsection{Limitations} 
We did not test our regularization scheme on stochastic differential equations. But it seems reasonable to expect that an extra care is needed to model such systems. The type of stochasticity may be effectively represented in latent space \cite{Li2020}, or it may resist a low-dimensional, time-invariant representation.

Another limitation of our model, and ODE models in general, would likely be encountered when attempting interpolation and forecasting performance for chaotic systems. While we have demonstrated good performance in coupled and non-periodic systems, finding robustness to out-of-distribution cases, the same strengths might lead to overconfident, temporally smooth predictions for systems that in reality behave chaotically. Care should therefore be taken to screen for and avoid systems in chaotic states.

\subsection{State-space modulation} 
Our path-minimizing loss encourages time invariance of the latents, but because the ODE solver operates on that space, some temporal evolution needs to be allowed. We expect that the time ``axis'' only occupies a low-dimensional submanifold, but our latents are not fully time-independent descriptions of the system configuration.
To further our goal of utilizing latent ODEs as robust inference models, we intend to combine our approach with the explicit static/dynamic state splitting of the latent parameters described by \cite{auzina2024modulated}.


\section{Conclusion}
\label{sec:conclude}
We introduce a novel regularization approach for latent ODE models. We remove the customary variational loss and replace it with an instance path-length penalty in latent space, akin to searching for geodesic paths on an implicitly defined data-driven manifold. This loss function can readily be used with all existing latent ODE architectures. 
We adopt the latent ODE-RNN architecture and find training with the proposed loss to be more effective, reaching lower loss values with fewer iterations. The resulting models significantly improve the interpolation and extrapolation accuracy in three different test cases. We also find increased robustness to parameter choices beyond the limits of the training data.

As intended, the latent distributions show little time evolution; instead, they are mostly shaped by system parameters and initial conditions.
The latent ODE encoder thus serves as an inference model, and we find accurate inference results when using the encoder to summarize irregularly sampled sequence data.
Our straightforward modification to the training loss thus manifestly improves the usability of latent ODE models to empirically describe dynamic systems.  

\section*{References}
\bibliographystyle{iopart-num}
\bibliography{refs} 

\clearpage
\appendix
\section{Experimental setup and ODE details}
\label{sec:AppendixA}

Our latent ODE-RNN architecture is implemented in \texttt{jax} and \texttt{equinox} \cite{jax2018github, kidger2021equinox}; the normalizing flow package is \texttt{flowjax} \cite{ward2023flowjax}.

\paragraph{ODE function}
We use a feed forward neural network to model our ODE function $f_{\theta}$ as in \cite{rubanova2019latent}. We use a Tanh activation function and report the network size for all trials in \autoref{sec:AppendixB}.

\paragraph{ODE solver}
For all data generation, and ODE integration of our models we use the ODE solver from the \texttt{diffrax} package \cite{kidger2021on}. We use the $5^{th}$ order \texttt{Tsit5()} solver, Tsitouras' 5/4 method ($5^{th}$ order Runge-Kutta), with adaptive steps and an initial $dt=0.1$.  We set the relative and absolute error tolerances of $1\mathrm{e}$-4 for both values. 

\subsection{Data generation}

We detail the data generation steps for the test cases below. We note we perform a $80,10,10$ split for training/testing/validation of all datasets. We add Gaussian noise to all generated data at a level of $0.05$ to encourage robustness.

\paragraph{Damped harmonic oscillator}
For this test we generate 10,000 samples of position and velocity data from numerically solving the equation
\begin{equation}
    m\frac{d^2x}{dt^2} = -kx - \frac{dx}{dt},
\end{equation}
where $k$ is the spring constant, and the mass ($m$) is set to unity.

Each sample consists of 30 data points that are sampled (uniform) randomly from $t_i=0$ to an endpoint of $t_f$, where we choose $t_f$ randomly from $[15, 20]$, to obtain irregularly spaced time intervals. We fix the spring constant to $k=0.12$ for all trials and uniformly sample for 2 initial conditions (position and velocity) from $[1,4]$.

\paragraph{Lane-Emden equation}
We generate 1,000 samples from a radius $\xi$, which we interpret as time, with 30 datapoints randomly sampled between $t_i = 0$ and $t_f \in [5, 9]$. To get the trajectories we numerically solve 
\begin{equation}
    \label{eqn:laneAppendix}
    \frac{1}{\xi^2} \frac{d}{d\xi} \left(\xi^2 \frac{d\theta}{d\xi} \right) + \theta^n = 0.
\end{equation}
As detailed in \autoref{sec:lane}, the initial condition for this problem is determined by physics, so only one degree of freedom remains: $n$. We uniformly sample $n \in \{1,2,3,4,5\}$ because values outside of this range, although physically possible, are numerically extremely unstable and thus studied far less frequently \cite{vanani2010numerical,baty2023modelling} (but see \autoref{sec:AppendixC} and \autoref{fig:lane2}). As the initial conditions are fixed, and we only sample 5 discrete values for $n$, the use of 1,000 trajectories is merely for training convenience to allow for batch sizes of 32.

\paragraph{Lotka-Volterra equations}
We generate 22,000 samples from numerical solutions to \autoref{eqn:lve}. Each sample consists of 150 data points are randomly sampled from $t_i=0$ to an endpoint of $t_f$, that we randomly choose from the $[25, 30]$.  As described in \autoref{sec:lve}, we uniformly sample for our parameters $\alpha \in [1, 3.5]$, $\beta\in [1, 3.5]$, $\gamma \in [0.5, 0.6]$, and $\delta \in [0.2, 0.3]$, and sample randomly for initial conditions between 1 and 6 for predator and prey numbers.

\section{Model and training details}
\label{sec:AppendixB}
All models are trained on a single Nvidia A100 GPU with 1 CPU and 40GB of memory. 

\paragraph{Damped harmonic oscillator}
Both latent ODE models presented (baseline, minimum path) consist of a hidden state of dimension 6, and a latent state of dimension 3. The ODE functions consist of 2 layers of 24 units with Tanh activations. We used the Adam optimizer with a learning rate of $1\mathrm{e}-2$ and trained for up to 10,000 steps with a batch size of 256. The best trained model was chosen for both the baseline and our model. We note we also varied the size of the hidden dimensions, latent dimension and neural ODE size between 2 and 16, 1 and 8, 16 and 40, respectively, and found no better results for either model.

For our path-minimized model we use a $\lambda$ of 1, finding good results for values anywhere in the range $[0.1, 5]$, with the main differences being the training times of the models.

\paragraph{Lane-Emden equation}
Both models presented are trained with a hidden state of dimension 6 and a latent space of dimension 3. The ODE functions consist of 2 layers of 24 units with Tanh activations. We used the Adam optimizer with a learning rate of $2\mathrm{e}$-3 and trained for up to 150,000 steps with a batch size of 32. We perform a variety of modifications to the model sizes and training hyperparameters in an attempt to allow the baseline to improve, however, we found no meaningful improvements even after going up to twice the size of hidden, latent, and ODE-RNN sizes and training for 120,000 steps (we see convergence with our model at $\sim 25,000$). We note the struggles of the baseline perhaps could be overcome with significantly larger ODE-RNN networks. But the purpose of this study was to compare the effectiveness of our regularization comparable models, hence did not push the baseline model past twice the size of our model.

We find the best results for $\lambda \in [0.1, 1]$, with the main differences again in training times.

\paragraph{Lotka-Volterra equations}
Both models presented are trained with a hidden state of dimension 16, and a latent state of dimension 8. The ODE functions consist of 3 layers of 40 units with Tanh activations. We used the Adam optimizer with a learning rate of $2\mathrm{e}$-3 and trained for up to 15,000 steps with a batch size of 64. The best trained model was chosen for both the baseline and our model. We also varied the size of the hidden dimensions, latent dimension and neural ODE size between 8 and 24, 4 and 12, 24 and 100, respectively, and found no better results for either model.

Previous studies \cite{shi2021segmenting, auzina2024modulated} report difficulties in training the baseline model to perform well on this problem, especially for extrapolation. They used complicated training schemes involving iterative growing scheme \cite{growing}, and/or sequentially increasing the size of the training time over multiple runs. We do not use any of these methods and simply perform a single training phase on a fixed training set.

We performed a large parameter sweep to find optimal values of $\lambda$ finding anything in the range of $\lambda \in [0.5, 2]$ to give us the best results and all reported results are from the trial with $\lambda = 0.5$.

\paragraph{Effect of penalty strength}
The $\lambda$ parameter in \autoref{eqn:loss} regulates the relative strength of the path minimization loss term compared to the reconstruction loss term. Optimal values for our three test cases we all within an order of magnitude of unity. To get a good first guess of $\lambda$, 
we try to equate early values of the reconstruction loss with the path minimization loss. This procedure requires minimal computation time and is allows for efficient hyperparameter searches.

We find training the LSTM models to be extremely difficult for this test case hence exclude the results as it does not provide an accurate view of the results.

\section{Further results}
\label{sec:AppendixC}

\subsection{Encoder DHO tests}
\begin{figure}
    \centering
    \includegraphics[width=0.95\linewidth]{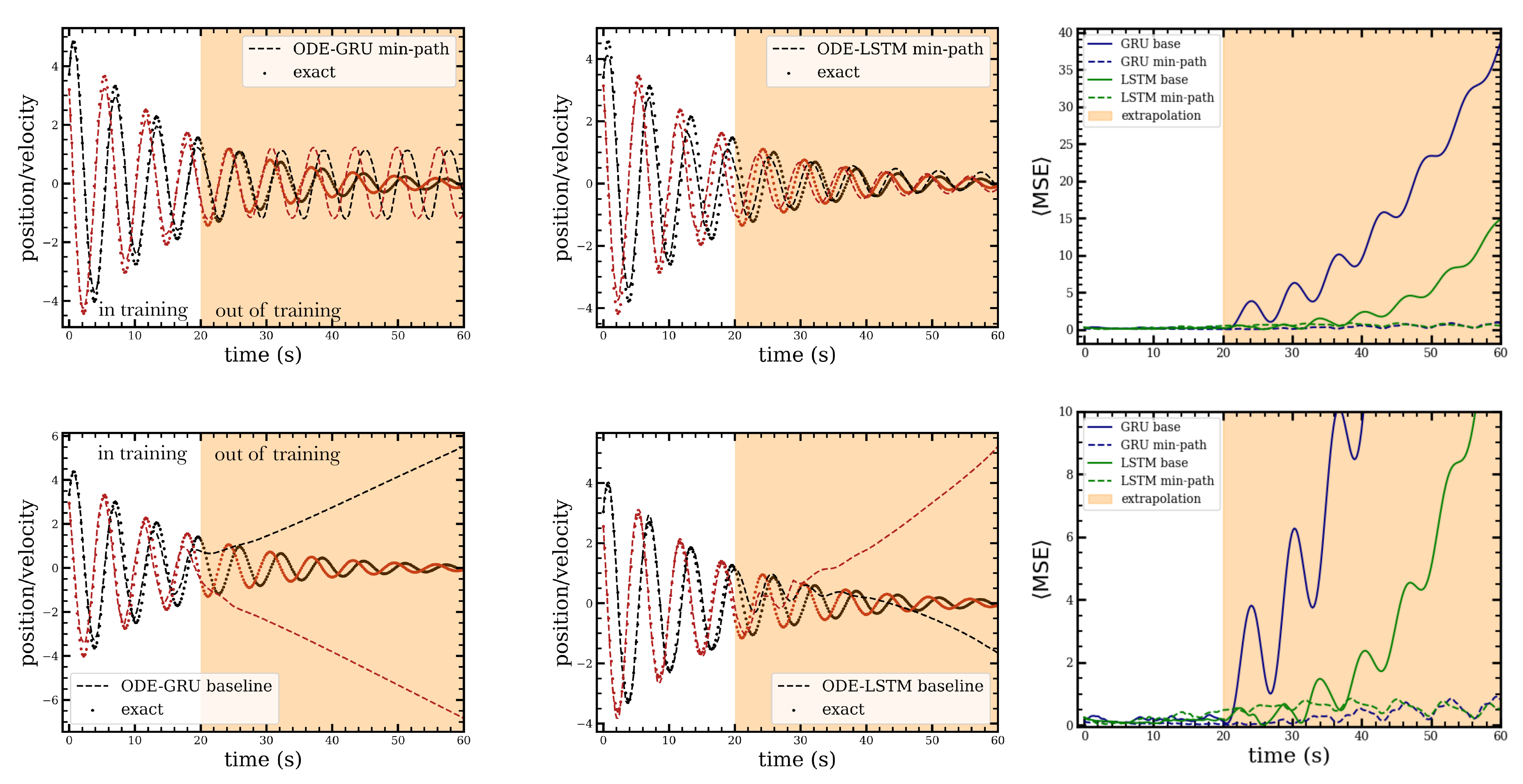}
    \caption{Panels showing the results of testing the path minimisation regularizer on the different encoders, with the baseline models in the bottom row and the path minimised in the top.}
    \label{fig:encoderTests}
\end{figure}

We show sample results from the damped harmonic oscillator test for ODE-GRU and ODE-LSTM encoder/decoder models in \autoref{fig:encoderTests} showing the MSE as a function of time in the rightmost panels.

\subsubsection{Out-of-domain spring constant}
\begin{figure}
    \centering
    \includegraphics[width=\linewidth]{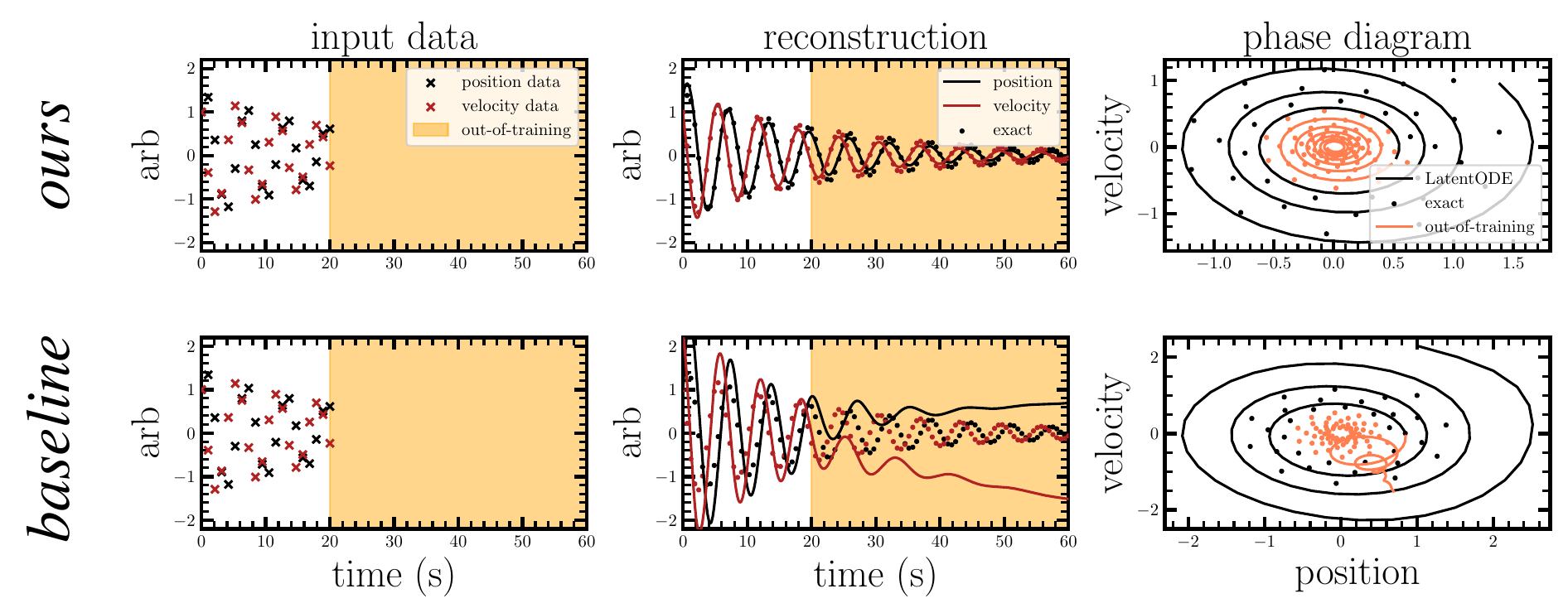}
    \caption{Same as \autoref{fig:sho_plot}, but for the spring constant 30\% lower than the training range.}
    \label{fig:dhok08}
\end{figure}

\begin{figure}
    \centering
    \includegraphics[width=\linewidth]{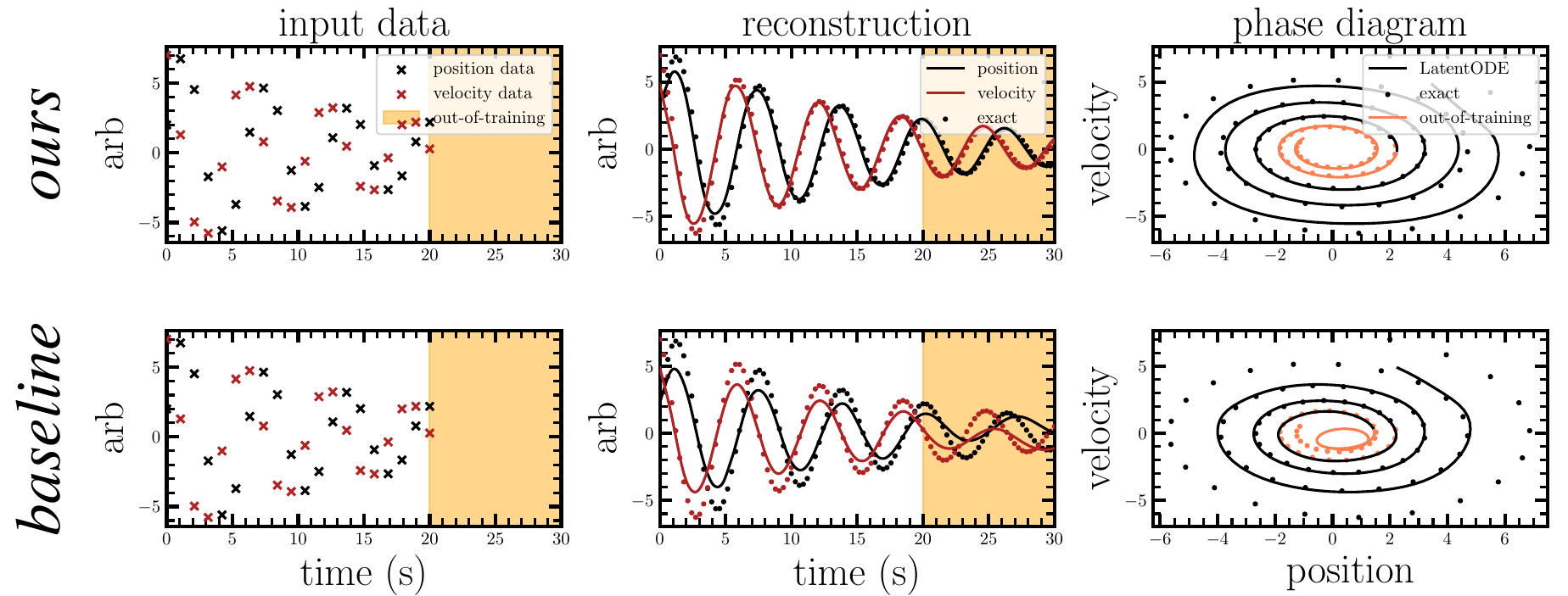}
    \caption{Same as \autoref{fig:sho_plot}, but with initial condition 30\% larger than the training range.}
    \label{fig:dhoIC}
\end{figure}

We tested baseline and out model on the damped harmonic oscillator by generating data where the spring constant $k$ was set $30\%$ lower than the training value (\autoref{fig:dhok08}) or with the ICs set $30\%$ higher than the training values (\autoref{fig:dhoIC}). This mirrors the out-of-distribution results presented in \autoref{tab:results}, with our model performing significantly better for interpolation and extrapolation than the baseline. We only show results for the best performing decoder model (ODE-RNN).

\subsubsection{Lane Emden: Non-integer polytropic indices}
\begin{figure}
    \centering
    \includegraphics[width=\linewidth]{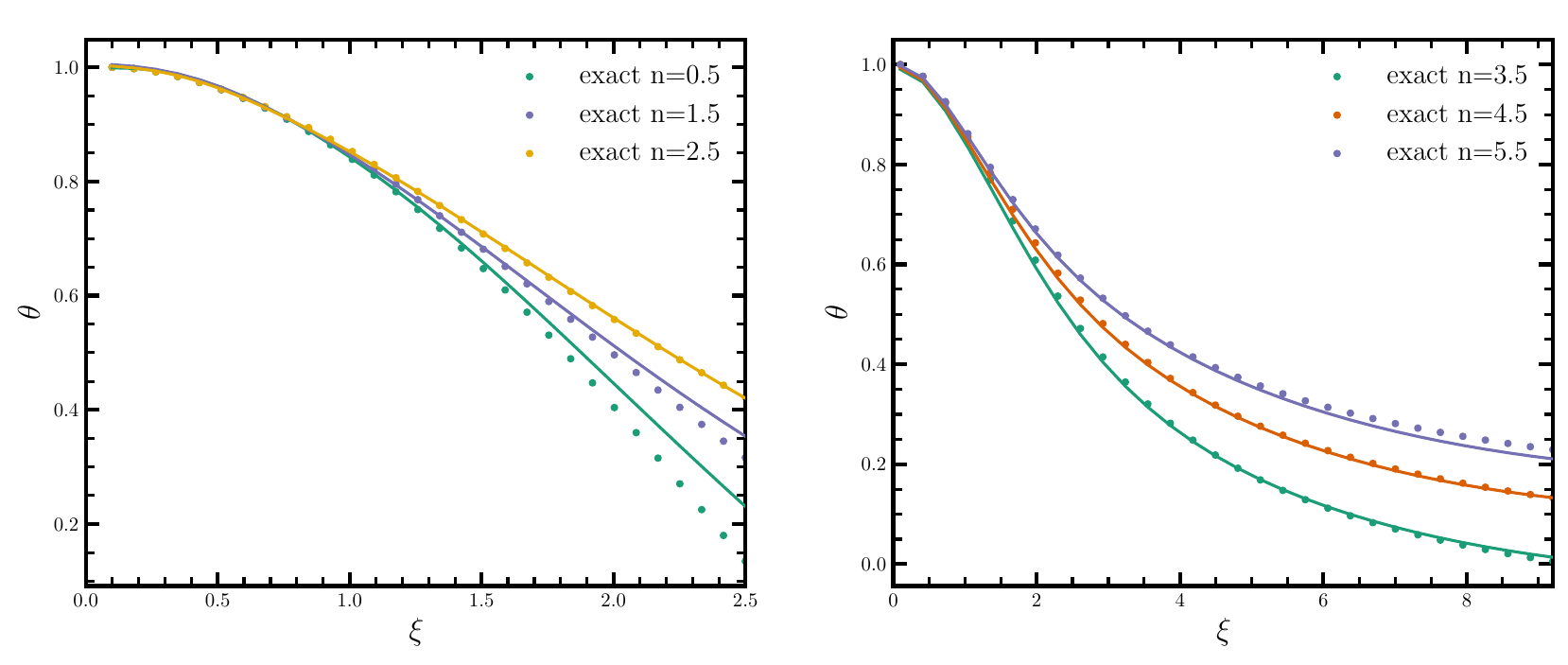}
    \caption{Non-integer solutions to the Lane Emden equation, solved numerically (points) and with our latent ODE model (lines). We show smaller non-integer indices in the left panel for a radius up to $\xi=3$, and larger values of $n$ plotted up to $\xi = 9$ in the right panel.}
    \label{fig:lane2}
\end{figure}

While we limit our training range for the Lane-Emden equation to integer values for the polytropic index $n$ in \autoref{eqn:lane}, we also tested results from our trained model for non-integer values of $n$ (\autoref{fig:lane2}). Due to the rapidly divergent nature of small non-integer values of $n$, we limit the radius $\xi$ for these values and plot them separately in the left panel. Because of the numerical instability, and the integer restriction during training, this tests shows a high level of robustness for out-of-distribution testing.

\subsection{Latent space structure in pred-prey system}
\begin{figure}
    \centering
    \includegraphics[width=0.8\linewidth, trim={10em 0 10em 0}]{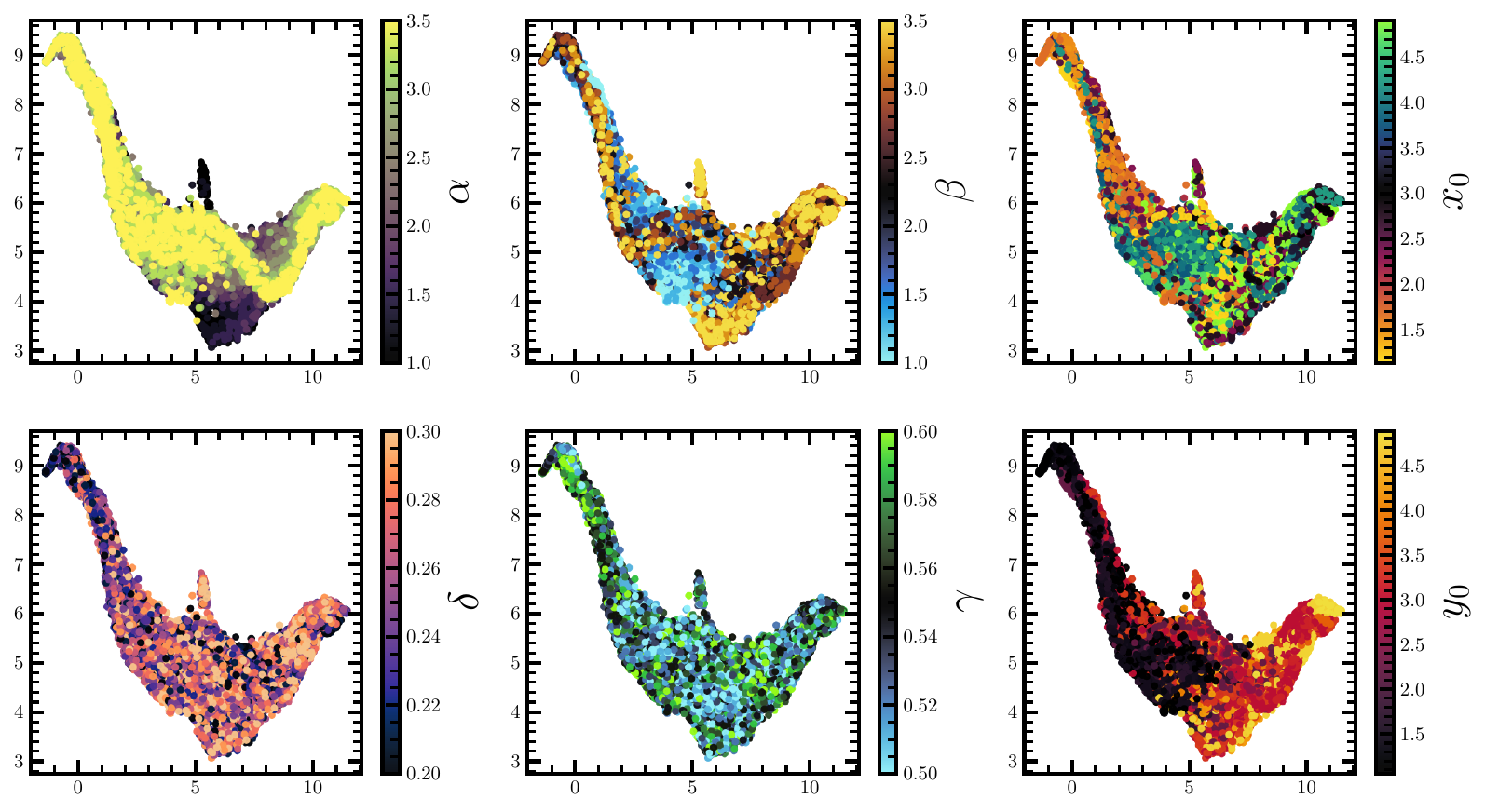}
    \caption{Same as the bottom panel of \autoref{fig:lve_umap}, except we also vary the initial conditions when constructing the UMAP showing them in the two rightmost panels.}
    \label{fig:umap2}
\end{figure}

\begin{figure}
    \centering
    \includegraphics[width=0.815\linewidth, trim={10em 0 10em 0}]{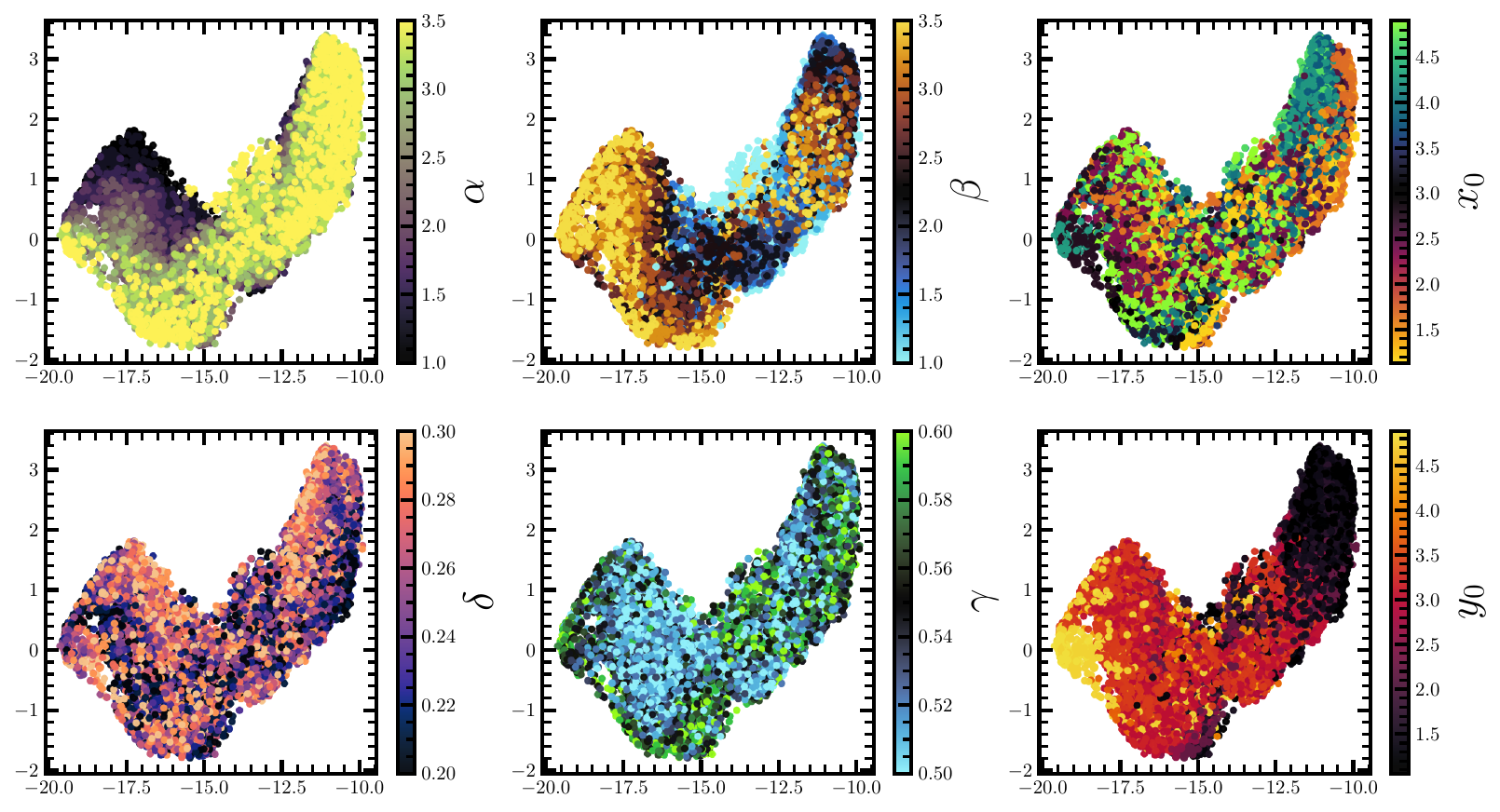}
    \caption{Same as \autoref{fig:umap2}, but for the baseline model.}
    \label{fig:umap3}
\end{figure}

We show additional UMAP projections of the latent space of both the baseline and our path-minimizing models for the predator-prey system in \autoref{fig:umap2} and \autoref{fig:umap3}. We show the parameters with free initial conditions, meaning we randomly sample the ICs over the 10,000 trials used to create the UMAP in all figures. This allows us to see that also the ICs are clearly located in latent space. Some of the trends are a little less clear than in \autoref{fig:lve_umap} because now six variable parameters are mapped onto two UMAP dimensions, but it is evident that the latents show sharper structures in our model than in the baseline, which is the reason for the improvements in the inference results.

\end{document}